\documentclass[11pt,letterpaper]{article}
\usepackage{emnlp2017}
\usepackage{times}
\usepackage{latexsym}
\usepackage{tikz}
\usepackage{listings}
\usepackage{algpseudocode}
\usepackage{algorithm}
\usetikzlibrary{positioning,calc,shapes,chains,trees,arrows,fit,plotmarks}
\usepackage{gnuplot-lua-tikz}
\emnlpfinalcopy

\def\emnlppaperid{***}

\title{Sparse Communication for Distributed Gradient Descent}

\author{}%Siddharth Patwardhan \and Preethi Raghavan \\
\author{Alham Fikri Aji \and Kenneth Heafield \\
        School of Informatics, University of Edinburgh \\
        10 Crichton Street\\
        Edinburgh EH8 9AB\\
        Scotland, European Union\\
        {\tt a.fikri@ed.ac.uk}, {\tt kheafiel@inf.ed.ac.uk}}

\date{}

\begin{document}

\maketitle

\begin{abstract}

We make distributed stochastic gradient descent faster by exchanging sparse updates instead of dense updates. Gradient updates are positively skewed as most updates are near zero, so we map the 99\% smallest updates (by absolute value) to zero then exchange sparse matrices. This method can be combined with quantization to further improve the compression. We explore different configurations and apply them to neural machine translation and MNIST image classification tasks. Most configurations work on MNIST, whereas different configurations reduce convergence rate on the more complex translation task. Our experiments show that we can achieve up to 49\% speed up on MNIST and 22\% on NMT without damaging the final accuracy or BLEU.

\end{abstract}

\section{Introduction}

Distributed computing is essential to train large neural networks on large data sets \cite{raina2009large}.  We focus on data parallelism: nodes jointly optimize the same model on different parts of the training data, exchanging gradients and parameters over the network.  This network communication is costly, so prior work developed two ways to approximately compress network traffic: 1-bit quantization \cite{1bitquant} and sending sparse matrices by dropping small updates \cite{strom2015scalable,dryden2016communication}.  These methods were developed and tested on speech recognition and toy MNIST systems.  In porting these approximations to neural machine translation (NMT) \cite{origneuralmt,bahdanau}, we find that translation is less tolerant to quantization.

Throughout this paper, we compare neural machine translation behavior with a toy MNIST system, chosen because prior work used a similar system \cite{dryden2016communication}. NMT parameters are dominated by three large embedding matrices: source language input, target language input, and target language output. These matrices deal with vocabulary words, only a small fraction of which are seen in a mini-batch, so we expect skewed gradients.  In contrast, MNIST systems exercise every parameter in every mini-batch. Additionally, NMT systems consist of multiple parameters with different scales and sizes, compared to MNIST's 3-layers network with uniform size.  More formally, gradient updates have positive skewness coefficient \cite{zwillinger1999crc}: most are close to zero but a few are large.

\section{Related Work}
An orthogonal line of work optimizes the SGD algorithm and communication pattern. \newcite{zinkevich2010parallelized} proposed an asynchronous architecture where each node can push and pull the model independently to avoid waiting for the slower node. \newcite{chilimbi2014project} and \newcite{NIPS2011_4390} suggest updating the model without a lock, allowing race conditions. Additionally, \newcite{dean2012large} run multiple minibatches before exchanging updates, reducing the communication cost.  Our work is a more continuous version, in which the most important updates are sent between minibatches. \newcite{DBLP:journals/corr/ZhangGLL15} downweight gradients based on stale parameters.

Approximate gradient compression is not a new idea. 1-Bit SGD \cite{1bitquant}, and later Quantization SGD \cite{DBLP:journals/corr/Alistarh0TV16}, work by converting the gradient update into a 1-bit matrix, thus reducing data communication significantly. \newcite{strom2015scalable} proposed threshold quantization, which only sends gradient updates that larger than a predefined constant threshold. However, the optimal threshold is not easy to choose and, moreover, it can change over time during optimization. \newcite{dryden2016communication} set the threshold so as to keep a constant number of gradients each iteration.  

\iffalse
\newcite{dryden2016communication} claimed that the approach worked well and can be combined with 1-bit SGD. However, their experiment were based on a small MNIST dataset. Thus, we will re-investigate this approach with a real dataset.

Other consideration is that \newcite{strom2015scalable} and \newcite{dryden2016communication} use a same threshold for the whole network. Where this might worked fine in small network such ash MNIST, we argue that this approach is non-optimal for larger or deeper network. Based on our investigation on the gradient updates in our NMT experiment, each network layer may have different skewness and different scale.

\fi

\section{Distributed SGD}

\begin{figure}[h!]
  \includegraphics[width=0.4\textwidth]{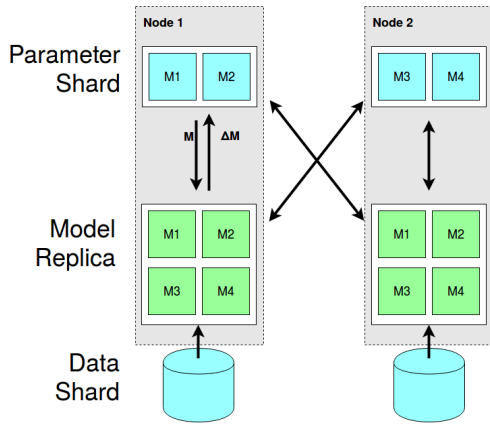}
  \caption{Distributed SGD architecture with parameter sharding.}
  \label{shard}
\end{figure}

We used distributed SGD with parameter sharding \cite{dean2012large}, shown in Figure~\ref{shard}. Each of the $N$ workers is both a client and a server.  Servers are responsible for $1/N$th of the parameters. 
%This architecture consists of $N$ workers where each worker holds a replica of the model and $1 / N$ parameters. 

Clients have a copy of all parameters, which they use to compute gradients.  These gradients are split into $N$ pieces and pushed to the appropriate servers.  Similarly, each client pulls parameters from all servers.  Each node converses with all $N$ nodes regarding $1/N$th of the parameters, so bandwidth per node is constant.

\section{Sparse Gradient Exchange}
\label{sec:graddrop}

We sparsify gradient updates by removing the R\% smallest gradients by absolute value, dubbing this Gradient Dropping. This approach is slightly different from \newcite{dryden2016communication} as we used a single threshold based on absolute value, instead of dropping the positive and negative gradients separately. This is simpler to execute and works just as well.  

Small gradients can accumulate over time and we find that zeroing them damages convergence.  Following \newcite{1bitquant}, we remember residuals (in our case dropped values) locally and add them to the next gradient, before dropping again.

\begin{algorithm}
\begin{algorithmic}
\Function{GradDrop}{$\nabla$, $R$}
    \State $\nabla += residuals$
    \State Select $threshold$: R\% of $|\nabla|$ is smaller
    \State $dropped \gets 0$
    \State $dropped[i] \gets \nabla[i] \forall i: |\nabla[i]| > threshold$
    \State $residuals \gets \nabla - dropped$
    \State \Return $sparse(dropped)$
\EndFunction
\end{algorithmic}
\caption{\label{drop_alg}Gradient dropping algorithm given gradient $\nabla$ and dropping rate $R$.}
\end{algorithm}

Gradient Dropping is shown in Algorithm~\ref{drop_alg}. This function is applied to all data transmissions, including parameter pulls encoded as deltas from the last version pulled by the client.  To compute these deltas, we store the last pulled copy server-side.  Synchronous SGD has one copy.  Asynchronous SGD has a copy per client, but the server is responsible for $1/N$th of the parameters for $N$ clients so memory is constant.  
%To achieve this, we have to store up to $N$ latest parameter histories over time. Each history requires $1/N$ memory, therefore all histories require $N * 1/N$ memory, which is constant to the number of nodes.

Selection to obtain the threshold is expensive \cite{alabi2012fast}. However, this can be approximated. We sample $0.1\%$ of the gradient and obtain the threshold by running selection on the samples. %By doing this, we can slightly reduce computation time.

Parameters and their gradients may not be on comparable scales across different parts of the neural network.  We can select a threshold locally to each matrix of parameters or globally for all parameters. In the experiments, we find that layer normalization \cite{2016arXiv160706450L} makes a global threshold work, so by default we use layer normalization with one global threshold.  Without layer normalization, a global threshold degrades convergence for NMT.  Prior work used global thresholds and sometimes column-wise quantization.  

\section{Experiment}
\label{sec:experiment}
We experiment with an image classification task based on MNIST dataset~\cite{lecun1998gradient} and Romanian$\rightarrow$English neural machine translation system.

For our image classification experiment, we build a fully connected neural network with three 4069-neuron hidden layers. We use AdaGrad with an initial learning rate of 0.005. The mini-batch size of 40 is used. This setup is identical to \newcite{dryden2016communication}.  

Our NMT experiment is based on \newcite{edinwmt16}, which won first place in the 2016 Workshop on Machine Translation.  It is based on an attentional encoder-decoder LSTM with 119M parameters. The default batch size is 80. We save and validate every 10000 steps. We pick 4 saved models with the highest validation BLEU and average them into the final model. AmuNMT~\cite{amun} is used for decoding with a beam size of 12. Our test system has PCI Express 3.0 x16 for each of 4 NVIDIA Pascal Titan Xs.  All experiments used asynchronous SGD, though our method applies to synchronous SGD as well.

\subsection{Drop Ratio}

To find an appropriate dropping ratio $R$\%, we tried 90\%, 99\%, and 99.9\% then measured performance in terms of loss and classification accuracy or translation quality approximated by BLEU~\cite{bleu} for image classification and NMT task respectively.

\begin{figure}[h!]
  \input{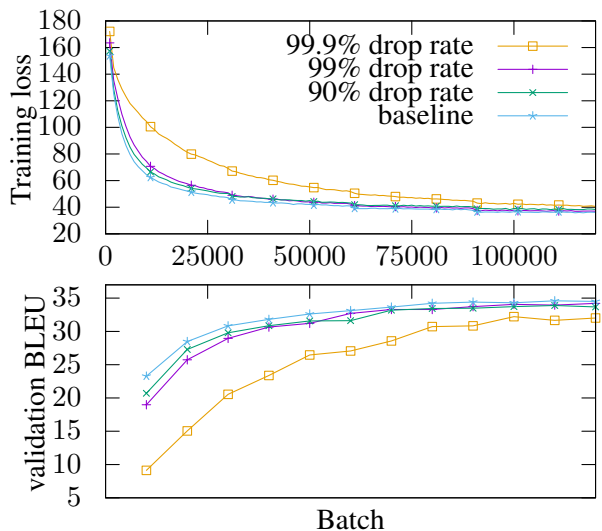}
  \caption{\label{fig:loss_to_ratio}NMT: Training loss and validation BLEU for different dropping ratios.}
  \end{figure}

Figure \ref{fig:loss_to_ratio} shows that the model still learns after dropping 99.9\% of the gradients, albeit with a worse BLEU score. However, dropping 99\% of the gradient has little impact on convergence or BLEU, despite exchanging 50x less data with offset-value encoding. The $x$-axis in both plots is batches, showing that we are not relying on speed improvement to compensate for convergence.

\newcite{dryden2016communication} used a fixed dropping ratio of 98.4\% without testing other options.   Switching to 99\% corresponds to more than a 1.5x reduction in network bandwidth.

For MNIST, gradient dropping oddly improves accuracy in early batches.  The same is not seen for NMT, so we caution against interpreting slight gains on MNIST as regularization.

  \begin{figure}[h!]
  \input{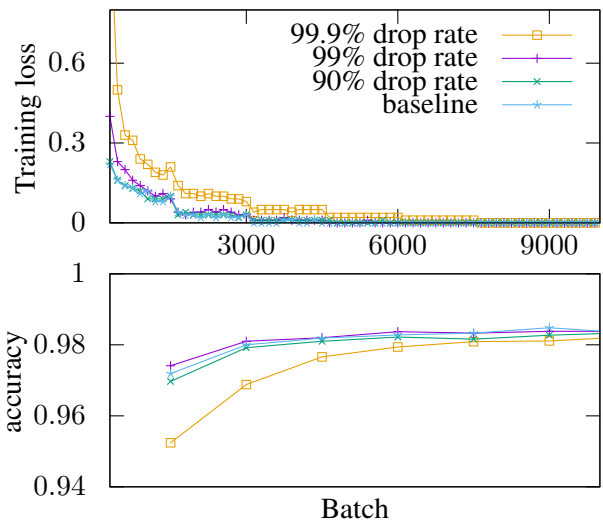}
  \caption{\label{fig:loss_to_ratio_MNIST}MNIST: Training loss and accuracy for different dropping ratios.}
  \end{figure}
  
\begin{table}
\begin{center}
\begin{tabular}{l r r}
\hline 
\textbf{Drop} & \textbf{words/sec} & \textbf{images/sec} \\
\textbf{Ratio} & \textbf{(NMT)} & \textbf{(MNIST)} \\
\hline
0\%& 13100 & 2489 \\  
90\% & 14443 & 3174 \\
99\% & 14740 & 3726 \\
99.9\% & 14786 & 3921 \\
\hline
\end{tabular}
\caption{Training speed with various drop ratios.}
\label{dropspeed}
\end{center}
\end{table}

\subsection{Local vs Global Threshold}
Parameters may not be on a comparable scale so, as discussed in Section~\ref{sec:graddrop}, we experiment with local thresholds for each matrix or a global threshold for all gradients. We also investigate the effect of layer normalization. We use a drop ratio of 99\% as suggested previously. Based on the results and due to the complicated interaction with sharding, we did not implement locally thresholded pulling, so only locally thresholded pushing is shown. 

\begin{figure}[h!]
  \input{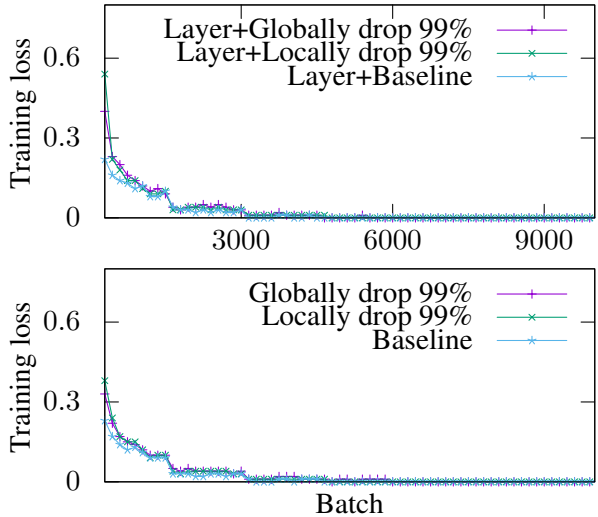}
  \caption{\label{normalized_MNIST}MNIST: Comparison of local and global thresholds with and without layer normalization.}
\end{figure}

\begin{figure}[h!]
  \input{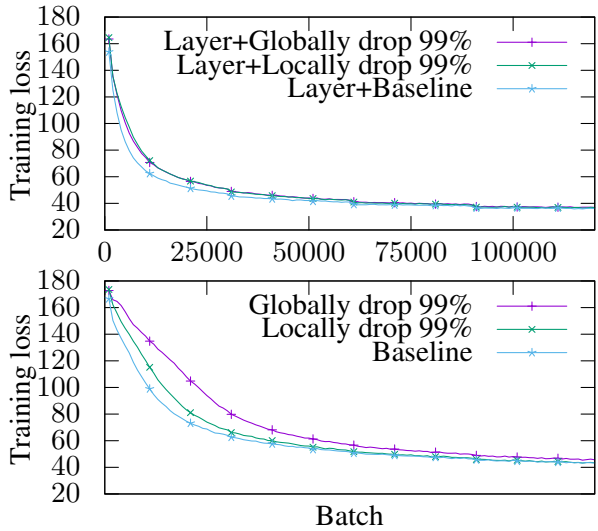}
  \caption{\label{normalized}NMT: Comparison of local and global thresholds with and without layer normalization.}
\end{figure}

The results show that layer normalization has no visible impact on MNIST. On the other side, our NMT system performed poorly as, without layer normalization, parameters are on various scales and global thresholding underperforms. Furthermore, our NMT system has more parameter categories compared to MNIST's 3-layer network.

\subsection{Convergence Rate}

While dropping gradients greatly reduces the communication cost, it is shown in Table \ref{dropspeed} that overall speed improvement is not significant for our NMT experiment. For our NMT experiment with 4 Titan Xs, communication time is only around 13\% of the total training time.  Dropping 99\% of the gradient leads to 11\% speed improvement. Additionally, we added an extra experiment of NMT with batch-size of 32 to give more communication cost ratio. In this scenario, communication is 17\% of the total training time and we see a 22\% average speed improvement. For MNIST, communication is 41\% of the total training time and we see a 49\% average speed improvement.  Computation got faster by reducing multitasking.

We investigate the convergence rate: the combination of loss and speed.  For MNIST, we train the model for 20 epochs as mentioned in \newcite{dryden2016communication}.  For NMT, we tested this with batch sizes of 80 and 32 and trained for 13.5 hours.

\begin{figure}[h!]
  \begin{tikzpicture}[gnuplot]
%% generated with GNUPLOT 5.0p6 (Lua 5.2; terminal rev. 99, script rev. 100)
%% Tue 20 Jun 21:14:46 2017
\path (0.000,0.000) rectangle (6.500,4.000);
\gpcolor{color=gp lt color border}
\gpsetlinetype{gp lt border}
\gpsetdashtype{gp dt solid}
\gpsetlinewidth{1.00}
\draw[gp path] (0.000,0.985)--(0.180,0.985);
\draw[gp path] (5.947,0.985)--(5.767,0.985);
\node[gp node right] at (-0.184,0.985) {$0.94$};
\draw[gp path] (0.000,1.867)--(0.180,1.867);
\draw[gp path] (5.947,1.867)--(5.767,1.867);
\node[gp node right] at (-0.184,1.867) {$0.96$};
\draw[gp path] (0.000,2.749)--(0.180,2.749);
\draw[gp path] (5.947,2.749)--(5.767,2.749);
\node[gp node right] at (-0.184,2.749) {$0.98$};
\draw[gp path] (0.000,3.631)--(0.180,3.631);
\draw[gp path] (5.947,3.631)--(5.767,3.631);
\node[gp node right] at (-0.184,3.631) {$1$};
\draw[gp path] (0.000,0.985)--(0.000,1.165);
\draw[gp path] (0.000,3.631)--(0.000,3.451);
\node[gp node center] at (0.000,0.677) {$0$};
\draw[gp path] (1.189,0.985)--(1.189,1.165);
\draw[gp path] (1.189,3.631)--(1.189,3.451);
\node[gp node center] at (1.189,0.677) {$100$};
\draw[gp path] (2.379,0.985)--(2.379,1.165);
\draw[gp path] (2.379,3.631)--(2.379,3.451);
\node[gp node center] at (2.379,0.677) {$200$};
\draw[gp path] (3.568,0.985)--(3.568,1.165);
\draw[gp path] (3.568,3.631)--(3.568,3.451);
\node[gp node center] at (3.568,0.677) {$300$};
\draw[gp path] (4.758,0.985)--(4.758,1.165);
\draw[gp path] (4.758,3.631)--(4.758,3.451);
\node[gp node center] at (4.758,0.677) {$400$};
\draw[gp path] (5.947,0.985)--(5.947,1.165);
\draw[gp path] (5.947,3.631)--(5.947,3.451);
\node[gp node center] at (5.947,0.677) {$500$};
\draw[gp path] (0.000,3.631)--(0.000,0.985)--(5.947,0.985)--(5.947,3.631)--cycle;
\node[gp node center,rotate=-270] at (-1.258,2.308) {Accuracy};
\node[gp node center] at (2.973,0.215) {time (second)};
\node[gp node right] at (4.479,1.627) {99\% drop rate};
\gpcolor{rgb color={0.000,0.620,0.451}}
\draw[gp path] (4.663,1.627)--(5.579,1.627);
\draw[gp path] (0.178,2.392)--(0.357,2.749)--(0.547,2.833)--(0.737,2.872)--(0.928,2.895)%
  --(1.106,2.961)--(1.296,2.899)--(1.487,2.912)--(1.689,2.903)--(1.891,2.921)--(2.081,2.921)%
  --(2.284,2.895)--(2.486,2.925)--(2.688,2.921)--(2.890,2.930)--(3.081,2.917)--(3.283,2.939)%
  --(3.485,2.917)--(3.687,2.943);
\gpsetpointsize{4.00}
\gppoint{gp mark 2}{(0.178,2.392)}
\gppoint{gp mark 2}{(0.357,2.749)}
\gppoint{gp mark 2}{(0.547,2.833)}
\gppoint{gp mark 2}{(0.737,2.872)}
\gppoint{gp mark 2}{(0.928,2.895)}
\gppoint{gp mark 2}{(1.106,2.961)}
\gppoint{gp mark 2}{(1.296,2.899)}
\gppoint{gp mark 2}{(1.487,2.912)}
\gppoint{gp mark 2}{(1.689,2.903)}
\gppoint{gp mark 2}{(1.891,2.921)}
\gppoint{gp mark 2}{(2.081,2.921)}
\gppoint{gp mark 2}{(2.284,2.895)}
\gppoint{gp mark 2}{(2.486,2.925)}
\gppoint{gp mark 2}{(2.688,2.921)}
\gppoint{gp mark 2}{(2.890,2.930)}
\gppoint{gp mark 2}{(3.081,2.917)}
\gppoint{gp mark 2}{(3.283,2.939)}
\gppoint{gp mark 2}{(3.485,2.917)}
\gppoint{gp mark 2}{(3.687,2.943)}
\gppoint{gp mark 2}{(5.121,1.627)}
\gpcolor{color=gp lt color border}
\node[gp node right] at (4.479,1.319) {baseline};
\gpcolor{rgb color={0.337,0.706,0.914}}
\draw[gp path] (4.663,1.319)--(5.579,1.319);
\draw[gp path] (0.274,2.608)--(0.571,2.762)--(0.880,2.877)--(1.178,2.877)--(1.475,2.868)%
  --(1.772,2.868)--(2.081,2.890)--(2.379,2.895)--(2.676,2.895)--(2.974,2.886)--(3.271,2.886)%
  --(3.568,2.872)--(3.866,2.872)--(4.163,2.868)--(4.472,2.872)--(4.769,2.886)--(5.067,2.881)%
  --(5.364,2.877)--(5.662,2.872);
\gppoint{gp mark 3}{(0.274,2.608)}
\gppoint{gp mark 3}{(0.571,2.762)}
\gppoint{gp mark 3}{(0.880,2.877)}
\gppoint{gp mark 3}{(1.178,2.877)}
\gppoint{gp mark 3}{(1.475,2.868)}
\gppoint{gp mark 3}{(1.772,2.868)}
\gppoint{gp mark 3}{(2.081,2.890)}
\gppoint{gp mark 3}{(2.379,2.895)}
\gppoint{gp mark 3}{(2.676,2.895)}
\gppoint{gp mark 3}{(2.974,2.886)}
\gppoint{gp mark 3}{(3.271,2.886)}
\gppoint{gp mark 3}{(3.568,2.872)}
\gppoint{gp mark 3}{(3.866,2.872)}
\gppoint{gp mark 3}{(4.163,2.868)}
\gppoint{gp mark 3}{(4.472,2.872)}
\gppoint{gp mark 3}{(4.769,2.886)}
\gppoint{gp mark 3}{(5.067,2.881)}
\gppoint{gp mark 3}{(5.364,2.877)}
\gppoint{gp mark 3}{(5.662,2.872)}
\gppoint{gp mark 3}{(5.121,1.319)}
\gpcolor{color=gp lt color border}
\draw[gp path] (0.000,3.631)--(0.000,0.985)--(5.947,0.985)--(5.947,3.631)--cycle;
%% coordinates of the plot area
\gpdefrectangularnode{gp plot 1}{\pgfpoint{0.000cm}{0.985cm}}{\pgfpoint{5.947cm}{3.631cm}}
\end{tikzpicture}
%% gnuplot variables
  \caption{MNIST classification accuracy over time.}
  \label{fig:full_acc}
\end{figure}
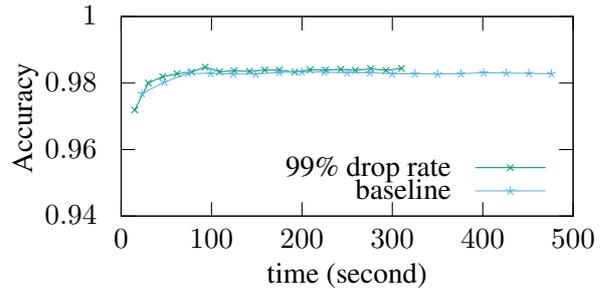

As shown in Figure \ref{fig:full_acc}, our baseline MNIST experiment reached 99.28\% final accuracy, and reached 99.42\% final accuracy with a 99\% drop rate. It also shown that it has better convergence rate in general with gradient dropping.

\begin{figure}[h!]
  \input{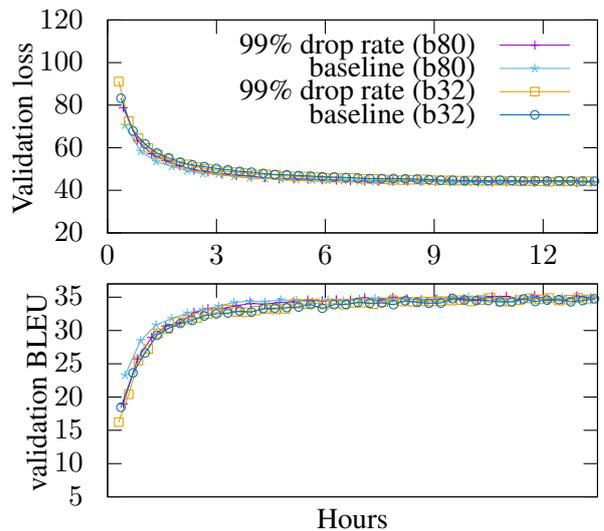}
  \caption{NMT validation BLEU and loss over time.}
  \label{fig:full_bleu}
\end{figure}

\begin{table}[!h]
\begin{center}
\begin{tabular}{ l c c }

\hline
\textbf{Experiment}&\textbf{Final}&\textbf{Time to reach}\\ 
 &\textbf{\%BLEU}&\textbf{33\% BLEU}\\ 
\hline
 batch-size 80 & & \\
 + baseline &  34.51 & 2.6 hours \\  
 + 99\% grad-drop  & 34.40 & 2.7 hours \\

 batch-size 32 & & \\  
 + baseline &  34.16 & 4.2 hours \\  
 + 99\% grad-drop  & 34.08 & 3.2 hours \\ 
\hline
\end{tabular}
\end{center}
  \caption{Summary of BLEU score obtained.}
  \label{summary}
\end{table}

Our NMT experiment result is shown in Table \ref{summary}. Final BLEU scores are essentially unchanged. Our algorithm converges 23\% faster than the baseline when the batch size is 32, and nearly the same with a batch size of 80.  This in a setting with fast communication: 15.75 GB/s theoretical over PCI express 3.0 x16. %This leads us to hypothesize that gradient dropping will be useful in multi-node scenarios, where communication is much more expensive.

\subsection{1-Bit Quantization}

We can obtain further compression by applying 1-bit quantization after gradient dropping.  \newcite{strom2015scalable} quantized simply by mapping all surviving values to the dropping threshold, effectively the minimum surviving absolute value.  \newcite{dryden2016communication} took the averages of values being quantized, as is more standard.  They also quantized at the column level, rather than choosing centers globally.  We tested 1-bit quantization with 3 different configurations: threshold, column-wise average, and global average. The quantization is applied after gradient dropping with a 99\% drop rate, layer normalization, and a global threshold.

\begin{figure}[h!]
  \input{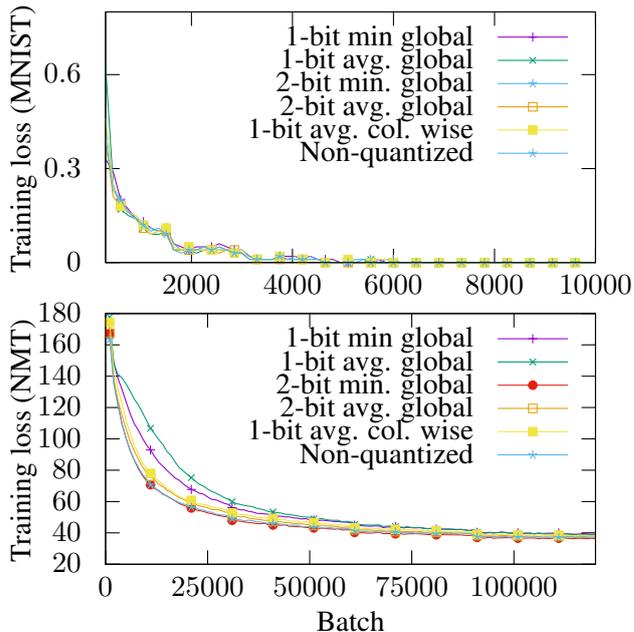}
  \caption{Training loss for different quantization methods.}
  \label{fig:quantized_result}
\end{figure}

Figure~\ref{fig:quantized_result} shows that 1-bit quantization slows down the convergence rate for NMT.  This differs from prior work \cite{1bitquant,dryden2016communication} which reported no impact from 1-bit quantization.  Yet, we agree with their experiments: all tested types of quantization work on MNIST.  This emphasizes the need for task variety in experiments.

NMT has more skew in its top 1\% gradients, so it makes sense that 1-bit quantization causes more loss.  2-bit quantization is sufficient.  

\section{Conclusion and Future Work}

Gradient updates are positively skewed: most are close to zero. This can be exploited by keeping 99\% of gradient updates locally, reducing communication size to 50x smaller with a coordinate-value encoding.

Prior work suggested that 1-bit quantization can be applied to further compress the communication. However, we found out that this is not true for NMT.  We attribute this to skew in the embedding layers. However, 2-bit quantization is likely to be sufficient, separating large movers from small changes.   Additionally, our NMT system consists of many parameters with different scales, thus layer normalization or using local threshold per-parameter is necessary. On the hand side, MNIST seems to work with any configurations we tried. 

Our experiment with 4 Titan Xs shows that on average only 17\% of the time is spent communicating (with batch size 32) and we achieve 22\% speed up. Our future work is to test this approach on systems with expensive communication cost, such as multi-node environments. 

\iffalse
Gradient updates are positively skewed: most are close to zero. This inspired us to sparsify the communication in distributed SGD by dropping the R\% smallest gradients (in absolute value) to zero. In NMT, dropping 99\% of the gradient reached a final BLEU of 34.4 compared to the baseline of 34.51 while reducing the communication size to 50x smaller with a coordinate-value encoding.

We found out that gradients in neural network model are mostly small numbers, with a few of them are significantly larger than the other. Thus, we proposed a way to force sparsity in gradient update by only keeping small portion of the gradients that have highest absolute value, and dropping the rest, and adding them to the next step's gradient. Our experiments show that the model is still capable of learning and producing a comparable final result even after dropping 99\% of the communication.

Theoretically, by dropping 99\% of the gradients every step, we can reduce the gradient's size to at least 50 times smaller, assuming we used a very naive coordinate+value sparse matrix encoding. The compression rate can be even higher with more efficient compression method. Therefore, we can see the potential of gradient drop to speed up distributed neural network learning process by reducing the communication time between nodes. In the future, we would like to try this method on larger distributed system of multiple machines and GPUs. Additionally, we would like to explore the usage of gradient drop in more general neural network system, such as image classification.

\fi

\section*{Acknowledgments}
Alham Fikri Aji is funded by the Indonesia Endowment Fund for Education scholarship scheme.
Marcin Junczys-Dowmunt wrote the baseline distributed code and consulted on this implementation.  
We thank School of Informatics computing and finance staff for working around NVIDIA's limit of two Pascal Titan Xs per customer.  Computing was funded by the Amazon Academic Research Awards program and by Microsoft's donation of Azure time to the Alan Turing Institute.  This work was supported by The Alan Turing Institute under the EPSRC grant EP/N510129/1.  

\bibliography{emnlp2017}
\bibliographystyle{emnlp_natbib}

\end{document}